\pdfoutput=1

\documentclass[11pt]{article}

\usepackage[final]{acl}

\usepackage{times}
\usepackage{latexsym}

\usepackage[T1]{fontenc}

\usepackage[utf8]{inputenc}

\usepackage{microtype}

\usepackage{inconsolata}

\usepackage{graphicx}

\usepackage{amsfonts}
\usepackage{amsmath}
\usepackage{xcolor}
\usepackage{multirow}
\usepackage{booktabs}
\usepackage{algorithm,algorithmic}

\usepackage{amssymb}
\usepackage{bbding}
\usepackage{listings}
\usepackage[most]{tcolorbox}
\usepackage{subcaption}
\usepackage{fontawesome}
\usepackage{tabularx}

\title{DCIS: Efficient Length Extrapolation of LLMs via Divide-and-Conquer Scaling Factor Search}

\author{%
  {\bf Lei Yang$^{1}$, Shaoyang Xu$^{2}$, Jianxiang Peng$^{1}$, Shaolin Zhu$^{1}$, Deyi Xiong$^{1,2}$\thanks{Corresponding author}}\\
  $^1$TJUNLP Lab, College of Intelligence and Computing, Tianjin University, Tianjin, China \\
  $^2$School of New Media and Communication, Tianjin University, Tianjin, China \\
  \texttt{\{yanglei\_9, zhushaolin, dyxiong\}@tju.edu.cn}
  }

\begin{document}
\maketitle

\begin{abstract}
Large language models (LLMs) based on the Transformer architecture usually have their context length limited due to the high training cost. Recent advancements extend the context window by adjusting the scaling factors of RoPE and fine-tuning. However, suboptimal initialization of these factors results in increased fine-tuning costs and reduced performance at target length. To address these challenges, we propose a novel RoPE-based fine-tuning framework that diverges from conventional scaling factors search. Specifically, we present a \textbf{D}ivide-and-\textbf{C}onquer \textbf{I}ncremental \textbf{S}earch (DCIS) algorithm that strategically determines the better scaling factors. Further fine-tuning with the identified scaling factors effectively extends the context window of LLMs. Empirical results demonstrate that our methodology not only mitigates performance decay at extended target lengths but also allows the model to fine-tune on short contexts and generalize to long contexts, thereby reducing the cost of fine-tuning. The scaling factors obtained through DCIS can even perform effectively without fine-tuning. Further analysis of the search space reveals that DCIS achieves twice the search efficiency compared to other methods. We also examine the impact of the non-strictly increasing scaling factors utilized in DCIS and evaluate the general capabilities of LLMs across various context lengths.
\end{abstract}

\begin{figure}[t]
    \centering
    \includegraphics[scale=0.43]{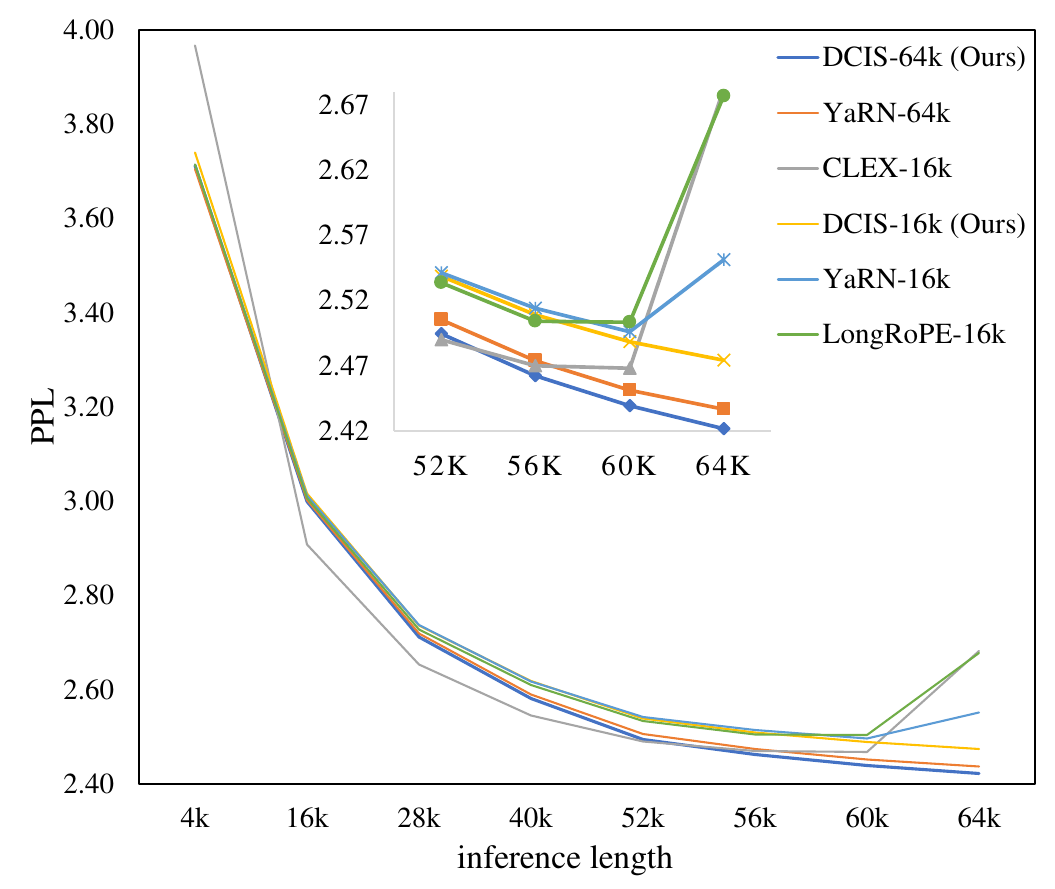}
    \caption{The Llama2-7B model is expanded to 64k context window. We test the PPL using 10 Proof-pile samples with a minimum length of 128k tokens. Fine-tuning is performed using the default method described in Section \ref{sec:exper}. $``\text{-64k/16k}"$ indicates that fine-tuning on a 64k/16k length generalizes to a target length of 64k.}
    \label{fig:overall}
\end{figure}

\section{Introduction}

Transformer \cite{vaswani2017attention} has emerged as the preferred architecture for large language models due to its scaling capability \cite{brown2020language, achiam2023gpt}. However, the inherent quadratic complexity of self-attention necessitates limiting the context window during pre-training, exemplified by the 4096-token limit in Llama2 \cite{touvron2023llama}. When models encounter sequences beyond this limit during inference, a significant loss in performance occurs \cite{press2021train, yang2025probench, guo2023evaluating}. To extend the operational context window of LLMs, previous studies have explored a wide variety of methods, such as sequence truncation \cite{rae2019compressive, dai2019transformer, wu2022memorizing} and sparse sequencing \cite{han2023lm, ding2023longnet, xiao2023efficient}, though these methods often result in the loss of critical contextual information.

Recent advances in positional encoding techniques have facilitated length generalization capabilities in LLMs, evolving from the initial sinusoidal positional encodings of Transformers to learnable and relative positional encodings \cite{gehring2017convolutional, shaw2018self, sun2024fuxitranyu}. The introduction of Rotary Position Embedding (RoPE) \cite{su2024roformer} has catalyzed a new wave of research that aims to increase the extrapolation length of LLMs by modifying the rotation frequency of the embedding dimensions through scaling factors \cite{peng2023yarn}, complemented by simultaneous fine-tuning to sustain long-context performance.

However, due to the quadratic complexity of self-attention and the increased number of intermediate activations cached during fine-tuning for long sequences \cite{shen2023large, pan2025advancing}, direct fine-tuning on target lengths \cite{peng2023yarn} leads to significantly high memory consumption and long time as the target sequence length increases. While fine-tuning on short contexts to generalize to target lengths \cite{chen2023clex} is more efficient, traditional approaches \cite{chen2023extending} often suffer from a significant performance drop \cite{chen2023clex, ding2024longrope} at the target length due to their limited generalization capability. As shown in Figures \ref{fig:overall} and \ref{fig:passkey_overall}, these existing methods fail to achieve satisfactory results in perplexity (PPL) and passkey performance metrics at target lengths.



To address these challenges, we explore better scaling factors to unlock the potential of LLMs for length generalization. 
We propose a novel RoPE-based fine-tuning framework that departs from traditional approaches of linearly increasing scaling factors \cite{peng2023yarn, ding2024longrope}. 
Instead, our framework introduces a Divide-and-Conquer Incremental Search (DCIS) algorithm, leveraging a principle of refinement from broad to specific, to efficiently determine the better scaling factors through continuous target-length inference. 
The non-strictly increasing nature of DCIS expands the search space considerably. 
The scaling factors identified via DCIS are then utilized for fine-tuning, significantly extending the model’s context window to the target length.

We conduct a comprehensive evaluation of DCIS on Llama2-7B, Llama3-8B and Mistral-7B-v0.1 using PPL and Passkey. Experimental results demonstrate that DCIS effectively mitigates the performance degradation of models at target context lengths.
Our contributions can be summarized as follows:
\begin{itemize}
    \item We propose a novel framework involving a Divide-and-Conquer Incremental Search (DCIS) algorithm for scaling factor search, followed by fine-tuning to extend the context window of LLMs.
    \item Extensive experiments demonstrate that DCIS overcomes the challenge of performance degradation at target lengths and exhibits strong generalization ability, leading to further reductions in fine-tuning costs. In addition, DCIS leads to substantial performance improvements even without fine-tuning.
    \item We conduct an in-depth analysis of DCIS, encompassing the impact of scaling factor initialization, the search space of DCIS, the role of Adaptive Scaling Factors (ASF), the effect of DCIS on general ability, sensitivity analysis of introduced hyperparameters, and scalability validation of the overall framework, demonstrating the method’s effectiveness, efficiency, and robustness.
\end{itemize}

\section{Related Work}

\textbf{Positional Embedding Scaling.} Recent advancements in positional embedding scaling, particularly involving Rotary Positional Embedding (RoPE), have significantly improved the capability of LLMs to manage extended context windows. Notable methods such as PI \cite{chen2023extending}, CodeLlama \cite{roziere2023code} and YaRN \cite{peng2023yarn} manually adjust scaling factors, whereas CLEX \cite{chen2023clex} optimizes rotation frequencies through training. LongRoPE \cite{ding2024longrope} employs a search mechanism to fine-tune scaling factors, enhancing the model's extrapolation abilities. While these approaches demonstrate improved performance through better scaling factors utilization, they are still hindered by high fine-tuning costs and notable performance declines at target lengths, issues that our proposed method addresses more efficiently.

\textbf{Sequence Compression.} Models leveraging RoPE have shown intrinsic capabilities for length extrapolation even without fine-tuning. Approaches like ReRoPE \cite{rerope2023} and Self-Extend \cite{jin2024llm} extend sequence lengths by compressing sequence indices, although they necessitate double attention computations, raising computational demands and limiting extrapolation potential. In contrast, our approach facilitates unrestricted extension of the model's context length through standard inference process, eliminating the need for repeated attention mechanisms.

\textbf{Chunking/Sparse Attention}. Investigative efforts have revealed that models predominantly focus on information at the sequence extremes \cite{liu2023lost}, suggesting that removing mid-sequence data while preserving initial and proximate tokens minimally impacts overall information integrity \cite{han2023lm, xiao2023efficient}. Alternative strategies involve segmenting sequences into chunks corresponding to pre-training lengths and employing external memory modules for storing and recalling past contexts during current chunk inference \cite{rae2019compressive, dai2019transformer, wu2022memorizing}. While these methods enable some degree of length extension, they inherently sacrifice a portion of the contextual data. Our method, however, maintains the integrity of the entire text, thereby maximizing the utility of the available context information.

\section{Preliminary}

This section elucidates the principles underpinning our approach and delineates the problem concerning positional embedding scaling, with a focus on enhancing the Rotary Positional Embedding (RoPE) \cite{su2024roformer} technique widely adopted in LLMs.

\subsection{Rotary Position Embedding}

RoPE has garnered significant attention in the realm of LLMs due to its robust performance and superior extrapolation effectiveness. Consider a sequence of embedding vectors $\mathbf{x}_1, \mathbf{x}_2, \dots, \mathbf{x}_L \in \mathbb{R}^{d}$, where $L$ represents the length of the input sequence and $d$ denotes the dimensionality of the hidden states for each head. Let $m$ indicate the position index. RoPE incorporates positional information into the embedding vectors through a rotational transformation defined as follows:
\begin{align} \label{ropeeq}
f_{\{Q,K\}}(\mathbf{x}_m, m, \mathbf{\Theta}) = \mathbf{R}^{d}_{\mathbf{\Theta}, m} \mathbf{W}_{\{Q, K\}} \mathbf{x}_m,
\end{align}
\noindent{where} $\mathbf{W}_Q$ and $\mathbf{W}_K$ denote the weights matrices of theTransformer. $\mathbf{R}^{d}_{\mathbf{\Theta}, m}$ is the rotation matrix parameterized by $\mathbf{\Theta}=\{\theta_i=10000^{-2(i-1)/d},i\in[1,2,\dots,d/2]\}$:
\begin{equation}
\begin{aligned}
\mathbf{R}^{d}_{\mathbf{\Theta}, m}&=\left( \begin{bmatrix}
\text{cos} m\theta_i & -\text{sin} m\theta_i \\
\text{sin} m\theta_i & \text{cos} m\theta_i 
\end{bmatrix} \right) ,\\
&\text{ for } i=1,2,\dots, \lfloor d/2\rfloor.
\end{aligned}
\end{equation}
This transformation ensures that the relative positional information $|m-n|$ is implicitly encoded in the attention scores:
\begin{equation}
\begin{aligned}
\mathbf{Q}^T_m \mathbf{K}_n &= (\mathbf{R}^{d}_{\mathbf{\Theta}, m} \mathbf{W}_Q \mathbf{x}_m)^T (\mathbf{R}^{d}_{\mathbf{\Theta}, n} \mathbf{W}_K \mathbf{x}_n) \\
 &= \mathbf{x}^T \mathbf{W}_Q \mathbf{R}^{d}_{\mathbf{\Theta}, n-m} \mathbf{W}_K \mathbf{x}_n.
\end{aligned}
\end{equation}

\subsection{Frequency Scaling}

\textbf{Frequency Scaling.} Despite RoPE's effectiveness in incorporating positional information, the model's capacity to handle sequences exceeding pre-training lengths remains limited, primarily due to inadequate training of low-frequency dimensions \cite{ntk} within the conventional context window size $L$. To address this, several scaling methods have been proposed to adjust RoPE's rotation frequency. We denote $\mathbf{R}^{d}_{\mathbf{\Theta}, m}$ succinctly as:
\begin{align} \label{dcis}
\left[\left(\text{cos} \left(\frac{m}{\lambda_i \beta_i} \right), \text{sin} \left(\frac{m}{\lambda_i \beta_i} \right)\right),i \in \left[0,\frac{d-2}{2} \right]\right],
\end{align}
where $\beta_i=10000^{2i/d}$ represents the base frequency, and $\lambda_i$ are the scaling factors for each frequency dimension.

\textbf{Scaling Factors Methods.} Both NTK-aware approach \cite{ntk} and YaRN \cite{peng2023yarn} apply theories from the Neural Tangent Kernel (NTK) and suggest varying the scaling factors $\lambda_i$ based on the dimension's training need, achieving better performance with less fine-tuning data. LongRoPE \cite{ding2024longrope}, on the other hand, utilizes a search-based method to identify better scaling factors $\lambda_i$, employing an evolutionary search algorithm for initial short text search (128k/256k) and fine-tuning, and search again at the target length (2M). This method facilitates up to a significant increase in processing length compared to the baseline, highlighting the potential of scaling adjustments in extending model capacities.

\begin{figure*}
    \centering
    \includegraphics[width=15.8cm]{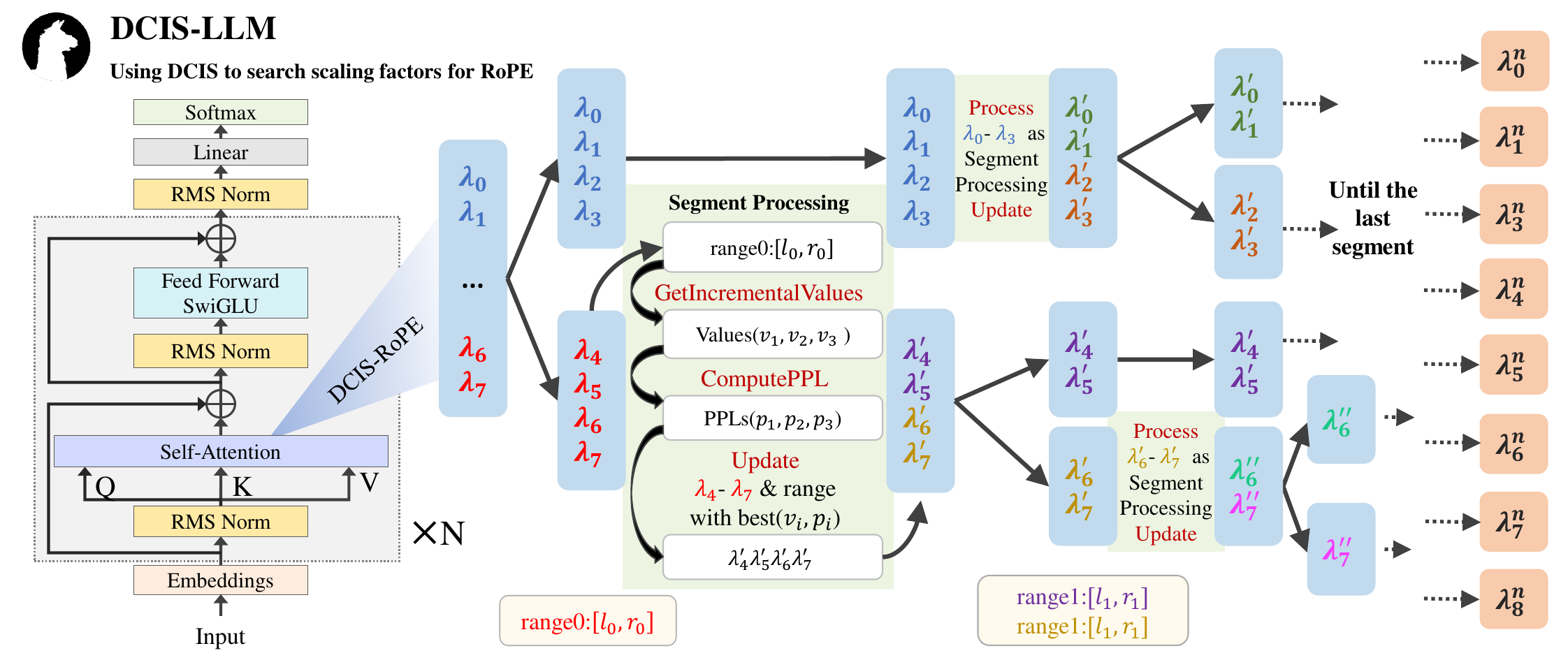}
    \caption{Diagram of the proposed DCIS framework. We illustrate the search procedure when $d=16$, with 3 incremental values selected for each processing step. Since the scaling factors are divided into high-frequency and low-frequency parts, we will initially process them in two segments. First, DCIS searches the scaling factors (i.e., $\lambda_4 - \lambda_7$) for the last 4 positions and gets 3 incremental values (i.e., $v_i$) within the range $[l_0,r_0]$. It then computes the PPL (i.e., $p_i$) for each incremental value. Finally, it selects the best incremental value with the lowest PPL to update these 4 scaling factors. As the input sequence is divided into two segments, DCIS processes the first 4 scaling factors in the first segment in the same manner. At the second layer, DCIS processes 2 scaling factors at a time, and at the third layer, it processes 1 scaling factor at a time, so on so forth. Finally, the process ends with the obtained scaling factors from the search.}
    \label{fig:algorithm}
\end{figure*}

\section{Methodology}

Although existing methods demonstrate generalization capabilities, they require significant fine-tuning data and suffer from marked performance degradation at extended target lengths. To address these challenges, we propose Divide-and-Conquer Incremental Search (DCIS) algorithm for length exploration. Figure \ref{fig:algorithm} illustrates our framework, beginning with an exploratory search phase using the Perplexity (PPL) metric as a guide, followed by fine-tuning using the identified scaling factors to effectively extend the model’s contextual modeling. In this section, we introduce our framework (Section \ref{sec:fra}), followed by a detailed description of the DCIS algorithm (Section \ref{sec:dcis}).

\subsection{Framework} \label{sec:fra}

\textbf{Searching Scaling Factors during Inference.} Recent research \cite{wu2024retrieval} reveals that specific retrieval heads contain long-form textual information, suggesting the intrinsic capacity of LLMs to process extended texts. To exploit this potential within LLMs, our framework first involves searching for optimal scaling factors during the preliminary inference phase. More specifically, our approach encourages the model to autonomously select suitable scaling factors through low-cost inference, thereby enhancing its extrapolation capability with low-cost training.

\textbf{Fine-Tuning with Searched Factors.} Directly applying the searched scaling factors to the original LLMs leads to suboptimal performance, indicating a lack of sufficient adaptation within the LLMs \cite{ding2024longrope}. Following YaRN, our framework further involves a fine-tuning phase with searched scaling factors. Due to our advanced search algorithm (Section \ref{sec:dcis}), better scaling factors are searched and initialized, thereby further reducing the number of fine-tuning steps.

\subsection{Divide-and-Conquer Incremental Search} \label{sec:dcis}

We elaborate our Divide-and-Conquer Incremental Search (DCIS) algorithm, which integrates the divide-and-conquer strategy to efficiently approximate better scaling factors. Beyond Figure \ref{fig:algorithm}, Algorithm \ref{alg:DCIS} presents the detailed procedure of DCIS. We also demonstrate that the search space of DCIS is half that of conventional search methods, in Section \ref{sec:searsp}.

\textbf{Algorithmic Strategy.} The scaling factors sequence, designated as $[\lambda_0, \lambda_1, \cdots, \lambda_{d/2-1}]$ (as formulated in Eq. \ref{dcis}), is methodically processed using a divide-and-conquer strategy. Specifically, we divide the sequence into a set of segments, and focus on one segment  (as highlighted in red in Figure \ref{fig:algorithm}) at a time while maintaining the others constant. For each segment, we adopt an incremental search strategy, where a new sequence of scaling factors is generated by adding a predetermined value from a set range to the current sequence. The most effective increment is then selected based on the lowest perplexity (PPL), which is also used to narrow the range for subsequent searches in the divided sub-segments. This iterative process, shown during the first iteration for $d=16$ with three values evaluated at each segment in Figure \ref{fig:algorithm}, incrementally refines the scaling factors.

As shown in Algorithm \ref{alg:DCIS}, the scaling factors are divided into high-frequency and low-frequency components. Hence we initially process them in two segments, with each segment handling $N= \text{head\_dim} / 2 $ scaling factors. In each iteration, the $\text{Segment}$ function is first employed to obtain the current segment $\text{Seg}$ that needs to be processed:
\begin{gather}
\text{Seg}=[(\lambda_{i},\ \lambda_{i+1},\cdots,\ \lambda_{i+N-1}), \\
\text{where}\ i=d-N\times j,\ j \in [1,\ d/N]].\notag
\end{gather}
\noindent{Subsequently}, the $\text{GetIncrementalValues}$ function is used to uniformly extract $C$ incremental values from the range of the current segment:
\begin{gather}
\text{Values}[k]=[R_{j,l}+\text{step}\times k], \\
\text{where}\ \text{step}=\frac{(R_{j,r}-R_{j,l})}{(C-1)},\ k\in [0,\ C-1].\notag
\end{gather}
\noindent{The} $\text{ComputePPL}$ function is then utilized to add these incremental values to the current scaling factors, yielding new scaling factors and thereby calculating $\text{PPLs}$:
\begin{gather}
\text{PPLs}(p_k)=\text{ComputePPL}(\text{Seg}+v_k).
\end{gather}
\noindent{Finally}, the $\text{PPLs}$ are used to update the scaling factors and the range of values for the next iteration. Specifically, the incremental value with the lowest PPL is used to update the scaling factors, while the lowest $C/3$ PPL incremental values are set as the range for the next iteration:
\begin{gather}
\text{PPLs, Values}=\text{sort}((p_k,v_k))\notag\\
\mathbf{F}_{\text{Seg}}=\text{Seg}+v'_1, \notag\\
R_{2\times j-1}=R_{2\times j}=[l,r], \\
\text{where}\ l=\text{min}(v'_1,v'_2,\cdots,v'_{C/3}), \notag\\
r=\text{max}(v'_1,v'_2,\cdots,v'_{C/3}).\notag
\end{gather}
This process continues until the scaling factors of the last segment are returned as the result of this search.

\begin{table*}[t]
    \centering
    \resizebox{\textwidth}{!}{
    \begin{tabular}{c c c c c c c c c c c c}
    \toprule
        \multirow{2}{*}{\textbf{Method}} & \multicolumn{2}{c}{\textbf{Fine-tuning}} & \multicolumn{9}{c}{\textbf{Inference Length}} \\
        & \textbf{Length} & \textbf{Data} & \textbf{4k} & \textbf{16k} & \textbf{28k} & \textbf{40k} & \textbf{52k} & \textbf{56k} & \textbf{60k} & \textbf{64k} & \textbf{AVG} \\
        \hline
        PI & 32k & - & 3.70 & 2.97 & 2.68 & 7.91 & 44.29 & 75.64 & 122.90 & 187.65 \textcolor{red}{$\uparrow$} & 55.97 \\
        \hline
        CodeLlama & 16k & 500B & 3.95 & 3.11 & 2.79 & 2.67 & 2.61 & 2.58 & 2.56 & 2.55 \textcolor{blue}{$\downarrow$} & 2.85 \\
        \hline
        \multirow{2}{*}{YaRN} & 64k & 0.8B & \underline{3.71} & 3.00 & 2.72 & 2.59 & 2.51 & 2.47 & \underline{2.45} & \underline{2.44} \textcolor{blue}{$\downarrow$} & \underline{2.74} \\
            & 16k  & 0.4B & 3.71 & 3.01 & 2.74 & 2.62 & 2.54 & 2.51 & 2.50 & 2.55 \textcolor{red}{$\uparrow$} & 2.77 \\
        \hline
        CLEX     & 16k     & 2.5B & 3.97 & \textbf{2.91} & \textbf{2.65} & \textbf{2.55} & \textbf{2.49} & \underline{2.47} & 2.47 & 2.68 \textcolor{red}{$\uparrow$} & 2.77 \\
        \hline
        LongRoPE   & 16k   & 0.4B & 3.72 & 3.01 & 2.73 & 2.61 & 2.53 & 2.50 & 2.50 & 2.68 \textcolor{red}{$\uparrow$} & 2.78 \\
        \hline
        \multirow{3}{*}{DCIS (Ours)} & 64k & 0.8B & 3.71 & 3.00 & \underline{2.71} & \underline{2.58} & \underline{2.49} & \textbf{2.46} & \textbf{2.44} & \textbf{2.42} \textcolor{blue}{$\downarrow$} & \textbf{2.73} \\
         & 16k & 0.4B & 3.74 & 3.02 & 2.74 & 2.62 & 2.54 & 2.51 & 2.49 & 2.47 \textcolor{blue}{$\downarrow$} & 2.77 \\
         & 4k & 0.8B & \textbf{3.68} & \underline{2.99} & 2.72 & 2.62 & 2.54 & 2.52 & 2.50 & 2.49 \textcolor{blue}{$\downarrow$} & 2.76 \\
        \hline
    \end{tabular}
    }
    \caption{PPL of different fine-tuning parameter configurations for various methods on Llama2-7B using the Proof-pile dataset. Blue arrows indicate a decrease in PPL at the target length, while red arrows indicate an increase.}
    \label{tab:main_ppl}
\end{table*}

\begin{figure*}[t]
    \centering
    \includegraphics[scale=0.56]{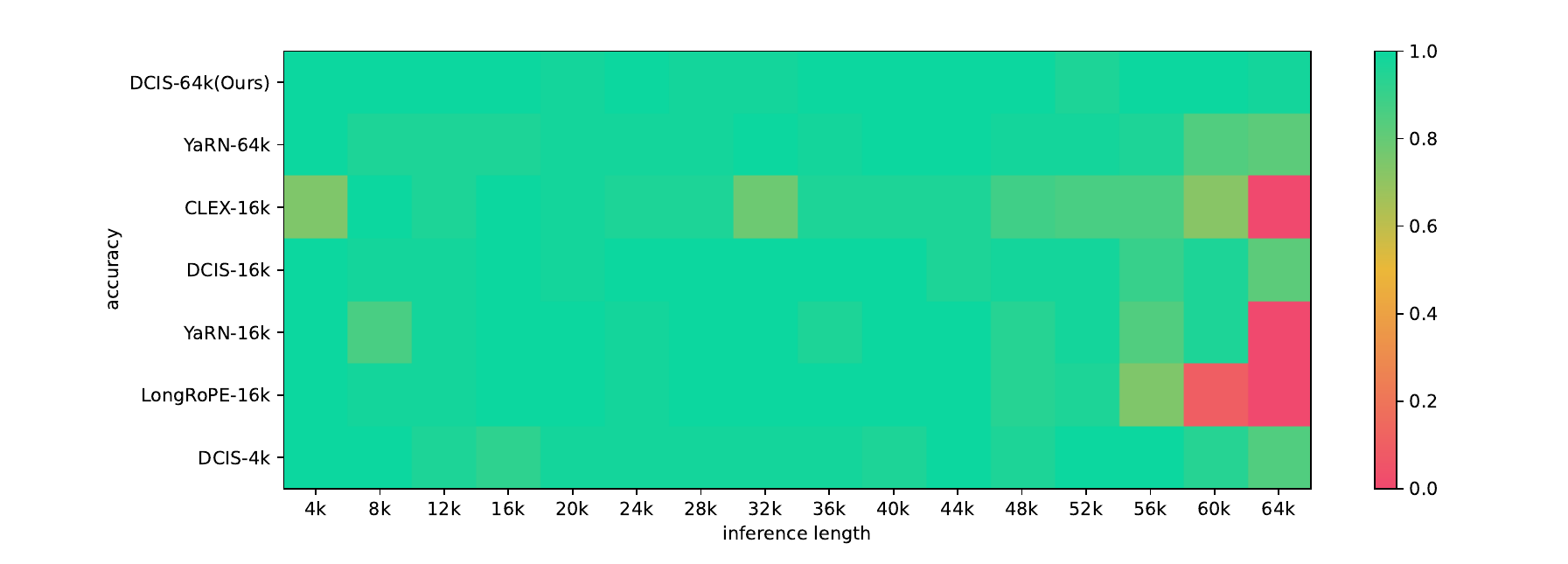}
    \caption{The recall rate of passkey with different lengths. Higher values indicate better performance.}
    \label{fig:passkey_overall}
\end{figure*}

\begin{figure*}[t]
    \centering
    \subfloat[Llama2/64k/Proof-pile]
    {
    \label{fig:l2_64}
    \includegraphics[width=4cm]{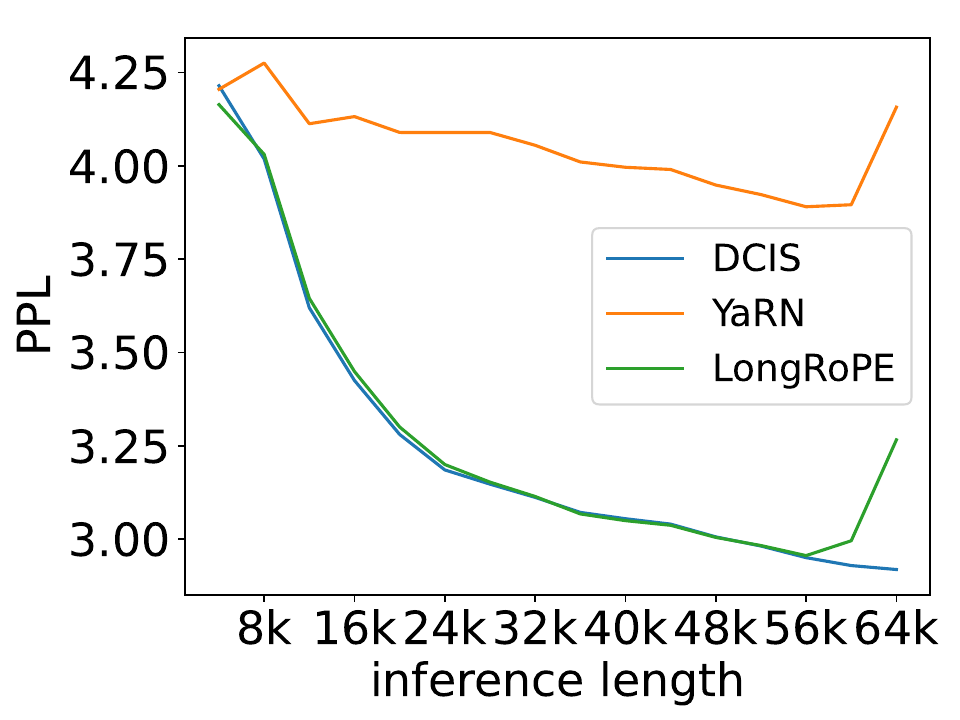}
    }
    \hspace{-0.5cm}
    \subfloat[Llama2/128k/Proof-pile]
    {
    \label{fig:l2_128}
    \includegraphics[width=4cm]{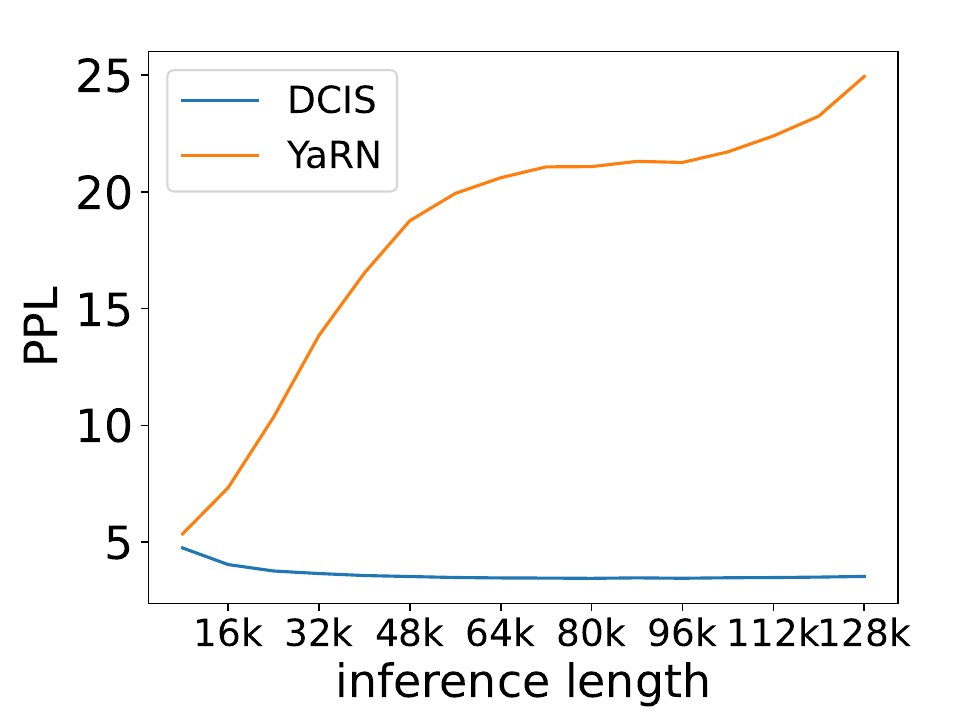}
    }
    \hspace{-0.6cm}
    \subfloat[Llama3/64k/PG-19]
    {
    \label{fig:l3_64}
    \includegraphics[width=4cm]{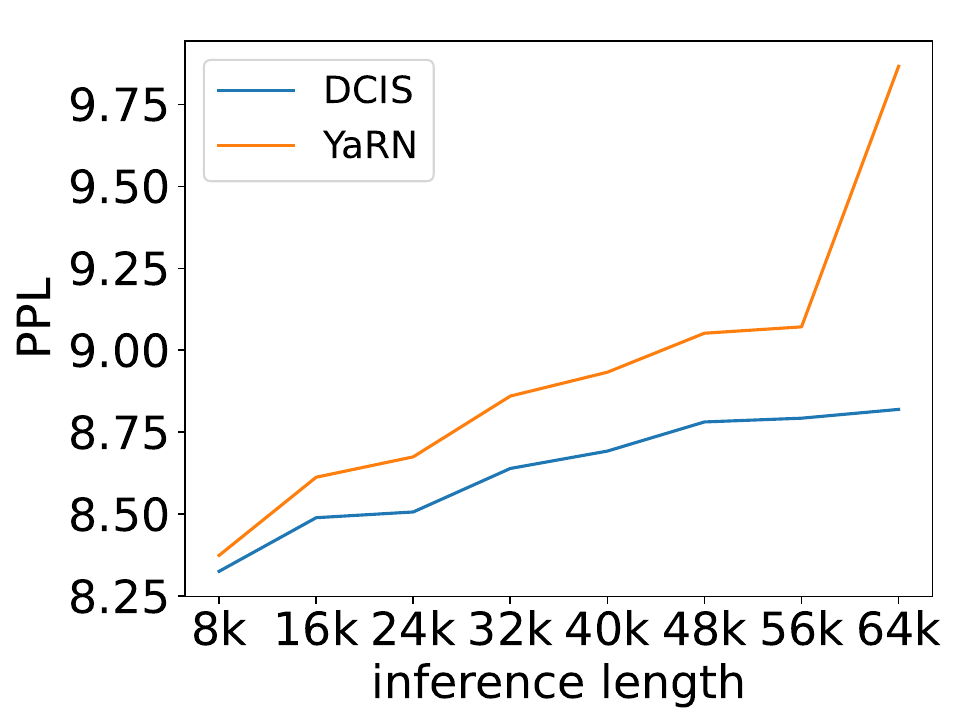}
    }
    \hspace{-0.47cm}
    \subfloat[Mistral/64k/PG-19]
    {
    \label{fig:ms_64}
    \includegraphics[width=4cm]{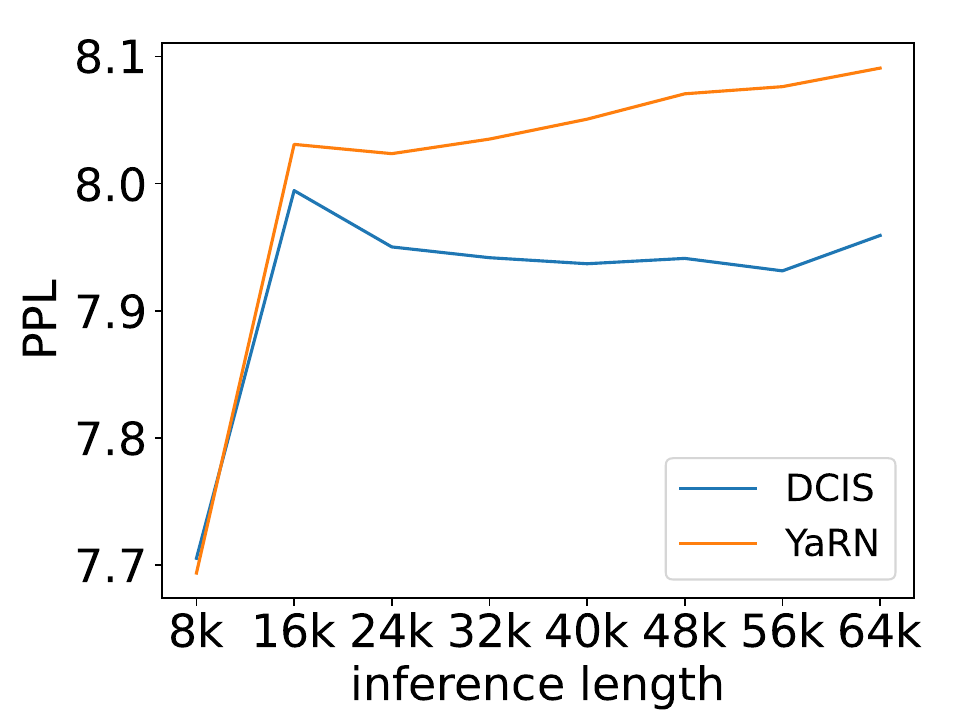}
    }
    \caption{PPL across different models, lengths, and datasets without fine-tuning.}
    \label{fig:noft}
\end{figure*}

\begin{figure*}[t]
    \centering
    \subfloat[Scaling factor distributions.]
    {
    \label{fig:fac}
    \includegraphics[width=6cm]{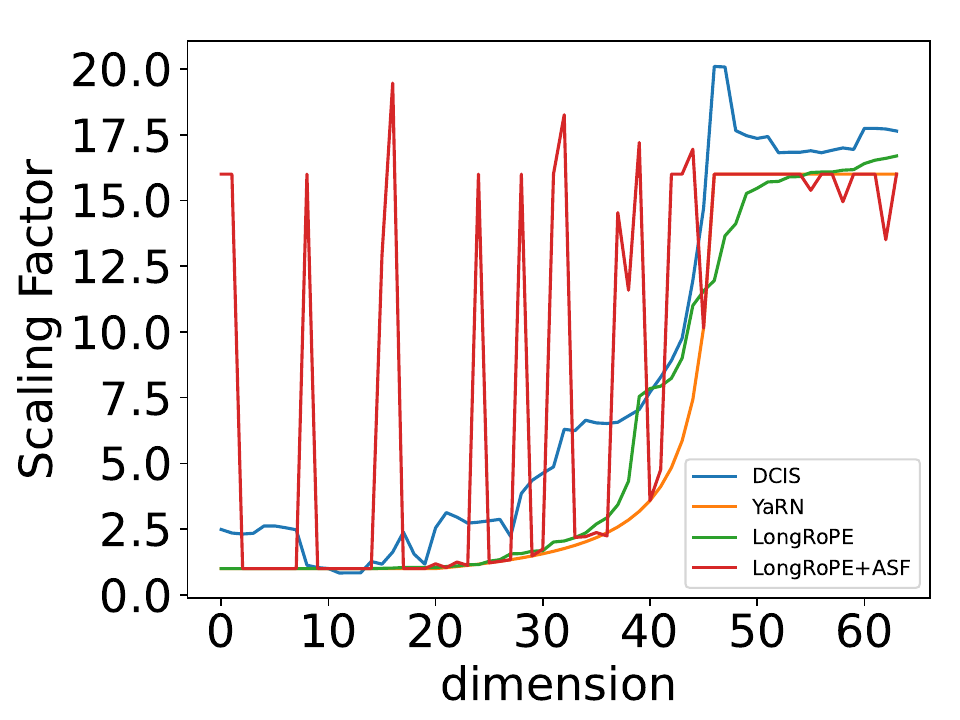}
    }
    \subfloat[PPL under two strategies.]
    {
    \label{fig:fac_pp}
    \includegraphics[width=6cm]{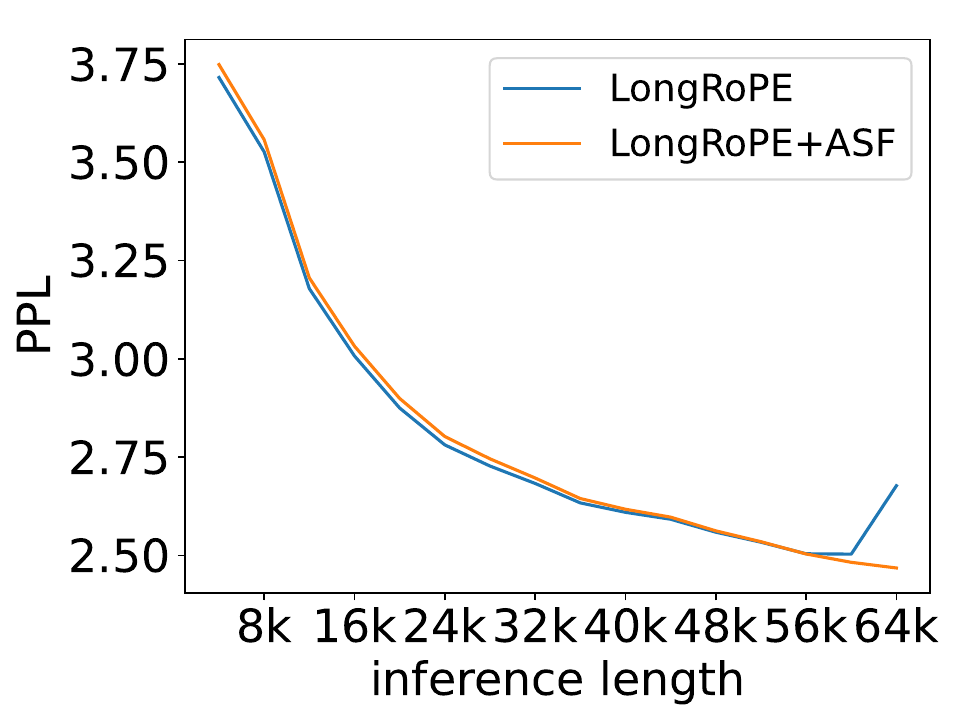}
    }
    \caption{Exploration of Adaptive Scaling Factor (ASF).}
    \label{fig:factors}
\end{figure*}

\begin{figure*}[t]
    \centering
    \includegraphics[scale=0.58]{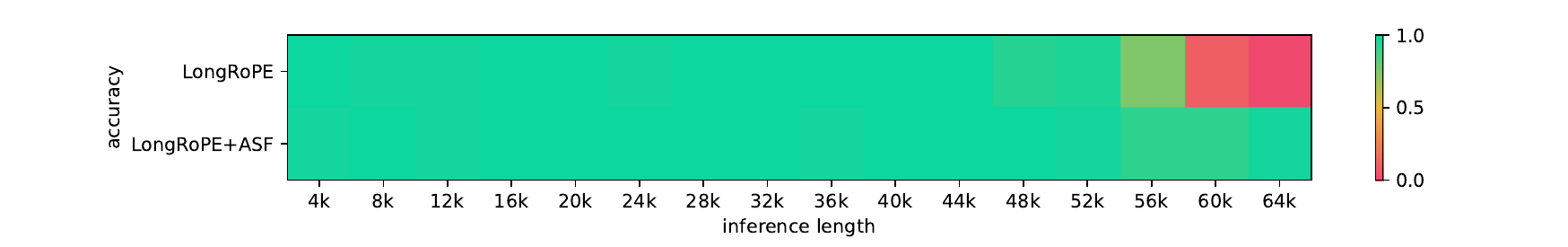}
    \caption{Recall rate under two strategies, the context length for fine-tuning is 16k.}
    \label{fig:factors_passkey}
\end{figure*}

\textbf{Optimizations Incorporated.} Based on empirical insights, several optimizations have been incorporated into the DCIS algorithm:
\begin{enumerate}
    \item Initial Scaling Factors: Drawing from the success of YaRN \cite{peng2023yarn}, its scaling factors are used as starting points from which our search begins.
    \item Priority to High-Dimensional Scaling Factors: Inspired by the NTK-aware approach \cite{ntk}, which suggests that higher dimensions might require more extensive interpolation, our algorithm prioritizes these dimensions for updates, thereby speeding up the convergence to better scaling factors.
    \item Discarding Non-Guiding Increments: Increments resulting in a PPL greater than 100 are considered ineffective and are thus excluded from the search to maintain focus on potentially successful modifications.
    \item Avoiding Local Optima: To prevent falling into local optima, when updating the range for the next layer of values, in addition to using the top $C/3$ PPL incremental values, we also expand the upper and lower bounds outward by one \text{step}. Here, one \text{step} is defined as the difference between adjacent incremental values.
\end{enumerate}

\begin{algorithm}[t]
	\small
	\caption{DCIS}
	\textbf{Input:} The target LLM, input samples $\mathbf{X}$,   initial scaling factors $\mathbf{F}$, initial range ${R}$, the number of increments processed each time $C$, the number of dimensions for each head $d$.\\
	\vspace{-2.5ex}

	\begin{algorithmic}[1]
		\label{alg:DCIS}
		\STATE $N$=$d/2$;
		\WHILE{$N$>=1}
        \FOR{$\text{Seg}$=\text{Segment}($N$)}
        \STATE \text{Values}=\text{GetIncrementalValues}($R$);
		\STATE \text{PPLs}=\text{ComputePPL}(LLM, $\mathbf{X}$, $\mathbf{F}$, $\text{Seg}$, \text{Values});
		\STATE  $\mathbf{F}$, ${R}$ = \text{update} ($\mathbf{F}$, ${R}$, \text{PPLs});
        \ENDFOR
		\STATE  $N$=$N/2$;
		\ENDWHILE
		\STATE Return the searched scaling factors $\mathbf{F}$;
	\end{algorithmic}
\end{algorithm}

\textbf{Adaptive Scaling Factors (ASF).} Unlike methods such as YaRN and LongRoPE, which prescribe a strictly increasing order for scaling factors as frequency decreases, our approach does not confine the model to predetermined scaling paths. Considering the complex and often opaque internal mechanisms of models, we posit that different dimensions within each head may require distinct treatments, some might even need interpolation despite being high-frequency components. Thus, our framework allows for flexible scaling factors adjustments, tailored to the specific needs of each dimension.

\section{Experiments} \label{sec:exper}

We conducted extensive experiments to examine the performance of the proposed DCIS across various metrics and conditions.


\subsection{Setup}

\textbf{Model and Evaluation Tasks}. We carried out our experiments using the Llama2-7B \cite{touvron2023llama}, Llama3-8B \cite{grattafiori2024llama} and Mistral-7B-v0.1 \cite{jiang2023mistral}, aiming to extend the model's context window to 64k tokens. We adopted YaRN's methodology for perplexity (PPL) evaluation, utilizing ten samples from the Proof-pile dataset \cite{rae2019compressive} with lengths of at least 128k tokens for assessment. Additionally, we assessed model performance using 50 passkey \cite{mohtashami2023landmark} tests at each length.

\textbf{Fine-tuning Parameters}. We used Llama2-7B as the base model. Initially, our DCIS algorithm was employed to identify scaling factors at the target length of 64k, setting the initial range between $[-5,5]$ with $C=10$ incremental values per segment. Following this, we segmented the PG19 dataset \cite{gao2020pile} into 4k, 16k, and 64k context lengths, and then performed fine-tuning on each segment. The fine-tuning process closely mirrors YaRN's protocol \cite{peng2023yarn} with a learning rate of $2 \times 10^{-5}$. For context lengths of \{4k, 16k, 64k\}, we employed total batch sizes of \{512, 64, 32\}, and all models were fine-tuned for 400 steps.

\textbf{Baseline}. By fine-tuning with the aforementioned hyperparameters, we obtained YaRN-\{16k, 64k\} \cite{peng2023yarn}, LongRoPE-16k \cite{ding2024longrope}, and our proposed DCIS-\{4k, 16k, 64k\} models. The CLEX-16k \cite{chen2023clex} used the original model. Additionally, to compare with other types of methods, we employed the InfLLM \cite{xiao2024infllm}, which claims to support infinitely long context windows by chunking and storing text sequences and retrieving the top-k most relevant chunks during inference.

\begin{table*}[t]
    \centering
    \begin{tabular}{cccccccc}
    \toprule
        \multirow{2}{*}{\textbf{Method}} & \textbf{Fine-tuning} & \multicolumn{5}{c}{\textbf{Categories}} \\
        & \textbf{Length} & \textbf{S-doc QA} & \textbf{M-doc QA} & \textbf{Sum} & \textbf{Few shot} & \textbf{Syn} & \textbf{Code} \\
        \hline
        YaRN 
        & 16k & 9.38 & 5.13 & 15.85 & 58.25 & 0.33 & 62.25 \\
        \hline
        CLEX & 16k & 6.93 & 8.30 & 13.63 & 57.68 & 0.69 & 44.06 \\
        \hline
        LongRoPE & 16k & 9.90 & 5.25 & 16.93 & 59.11 & 0.31 & 61.49 \\
        \hline
        InfLLM & - & 6.40 & 5.10 & 6.17 & 51.05 & 0.78 & 62.27 \\
        \hline
        \multirow{2}{*}{DCIS(Ours)} 
        & 16k & 7.58 & 3.51 & 16.11 & 59.10 & 0.42 & 62.05 \\
        & 4k & 8.17 & 4.09 & 14.55 & 58.07 & 0.25 & 61.67 \\
        \hline
    \end{tabular}
    \caption{Evaluation of different methods on the LongBench benchmark.}
    \label{tab:longbench}
\end{table*}

\subsection{Main Results} \label{sec:mainrs}

We assessed the PPL of our proposed model and baseline models on the Proof-pile dataset, with results shown in Figure \ref{fig:overall} and Table \ref{tab:main_ppl}. Figure \ref{fig:passkey_overall} depicts the performance on the passkey evaluation. Notably, models from other methods that were fine-tuned on 16k-length and subsequently generalized to a 64k context windows experienced a marked increase in PPL at the target length (64k). Furthermore, these models entirely failed the passkey test at the target length. In stark contrast, our approach exhibited consistent performance across various context lengths, including the target length, and even outperformed other methods at shorter fine-tuning lengths (4k). Additionally, when fine-tuned on a 64k-length context, DCIS consistently outperforms YaRN on both PPL and passkey metrics.

All results underscore the significance of scaling factors. DCIS identify superior scaling factors, leading to improved performance across various sequence lengths and demonstrating strong generalization capabilities, thereby reducing the memory, data, and time associated with model fine-tuning. The superior initial scaling factors contribute to further reduction in the number of fine-tuning steps. For example, in Appendix \ref{apd:fts}, we report our observations that our method requires fewer fine-tuning steps to achieve comparable performance to YaRN.

\subsection{DCIS without Fine-Tuning} \label{sec:woft}
Figure \ref{fig:noft} illustrates the outcomes of our experiments wherein inference was conducted solely by adjusting scaling factors without fine-tuning. A comprehensive search for scaling factors was performed at target lengths of 64k and 128k, followed by direct PPL evaluation. Figures \ref{fig:l2_64} and \ref{fig:l2_128}, clearly indicate a consistent decline in our model's PPL values at both target lengths, in stark contrast to the upward trends observed in other methods. Furthermore, we extended our experiments to the newly released LLama3-8B and the different architecture, Mistral, employing the DCIS algorithm for scaling factors search, and evaluated the resulting models on the PG19 \cite{raecompressive2019} test set. Figures \ref{fig:l3_64} and \ref{fig:ms_64}, demonstrate that our approach consistently achieves the lowest PPL across all settings, further validating its broad applicability and effectiveness.


In Appendix \ref{apd:slppl}, we compare the PPL of scaling factors for various methods at shorter lengths, without fine-tuning.

\section{Analysis}

In this section, we delved into a comprehensive analysis of DCIS. We began by contrasting its search space with other search methods. Subsequently, we examined the implications of non-strictly increasing scaling factors. Furthermore, to evaluate the model's general ability on real-world tasks, we employed LongBench \cite{bai2023longbench} and Open LLM Leaderboard from Hugging Face to assess the model's performance on long and short contexts. Finally, we conducted an in-depth investigation into the sensitivity analysis of hyperparameters introduced by the DCIS method, as well as the scalability performance of the proposed framework across different scenarios.

\subsection{Search Space Analysis} \label{sec:searsp}

The efficiency of our DCIS algorithm is highlighted by comparing the search space. Specifically, our search space is the product of the total number of processed segments $d-2$ and the number of increments per segment $C$. The search space of evolutionary search utilized by LongRoPE is the product of the number of iterations $T$ and the population size $P$. For example, for the Llama2-7B model with default parameters, our search space equates to $(d-2) \times C = (128-2) \times 10 = 1260$. In contrast, the search space of evolutionary search calculates as $T \times P = 40 \times 64 = 2560$, signifying that our algorithm's search speed is effectively double that of evolutionary search.

\subsection{Non-Strictly Increasing Scaling Factor}  \label{sec:nostr}

Since LongRoPE utilizes strictly monotonically increasing scaling factors, we performed ablation studies on it to examine the effects of our proposed ASF. 
Specifically, we adjusted the evolutionary search algorithm in LongRoPE to a non-strictly increasing one, applied it to fine-tune the model on a 16k-length context, and subsequently generalized it to a 64k context window. Figures \ref{fig:factors} and \ref{fig:factors_passkey} depict the experimental outcomes. 
Figure \ref{fig:fac} visualizes the scaling factor distributions employed by various methods. 
Notably, both DCIS and LongRoPE + ASF exhibited irregular, sawtooth-like scaling factors, while YaRN and the original LongRoPE demonstrated more stable scaling factors. 
Furthermore, as shown in Figures \ref{fig:fac_pp} and \ref{fig:factors_passkey}, our ASF can improve LongRoPE in terms of both PPL and passkey scores, suggesting that imposing fewer constraints on scaling factors may enhance model performance.




\begin{table}[t]
    \centering
    \resizebox{\columnwidth}{!}{
    \begin{tabular}{c c c c c c}
    \toprule
        $[l,r]$ & $[-3, 3]$ & $[-4, 4]$ & $[-5,5]$ & $[-6, 6]$ & $[-7, 7]$ \\
        \hline
        PPL & 10.2 & 10.2 & 10.2 & 10.2 & 10.2 \\
        \hline
    \end{tabular}
    }
    \caption{PPL for different initial ranges $[l,r]$ with fixed $C=10$.}
    \label{tab:init_range}
\end{table}

\begin{table}[t]
    \centering
    \begin{tabular}{c c c c c}
    \toprule
        $C$ & 6 & 8 & 10 & 12 \\
        \hline
        PPL & 10.4 & 10.2 & 10.2 & 10.2 \\
        \hline
    \end{tabular}
    \caption{PPL for different $C$ with fixed $[l,r]=[-5,5]$.}
    \label{tab:samp_cnt}
\end{table}

\subsection{General Ability Evaluation} 
\label{sec:gae}

In addition to the previous assessments on PPL and passkey, employing LongBench and Open LLM Leaderboard, we conducted a comprehensive assessment of the models' general abilities in both long and short context scenarios. The empirical results, as depicted in Tables \ref{tab:longbench} and \ref{tab:benchmarks}, indicate that there is no significant difference in performance among the various methods, with different models performing best on different subsets. While models utilizing full attention generally achieved slightly better results, InfLLM, which leverages a retrieval-based approach, demonstrated a notable advantage in terms of memory efficiency. These findings suggest that the optimal choice of model is contingent upon specific application requirements.

\subsection{Hyperparameter Sensitivity}
The DCIS algorithm introduces two critical hyperparameters: the initial search range $[l,r]$ and the number of increments per segment $C$.

Our experimental observations demonstrate that the DCIS algorithm exhibits the following characteristics during execution. 1) Coarse-grained iteration phase. In the initial iterations, the PPL values obtained from each sampling display a U-shaped distribution, indicating that the initial range $[l,r]$ already encompasses the optimal value region. 2) Fine-grained iteration phase. In subsequent refinement searches, PPL variations tend to stabilize, suggesting that the algorithm has converged to the vicinity of the optimal value.

Based on these observations, we can derive the following theoretical analysis. 1) Initial range $[l,r]$. As long as this range covers the region near the optimal value, its specific magnitude has a limited impact on the final results, rendering the algorithm insensitive to this parameter. 2) The number of increments per segment $C$. Theoretically, larger values are preferable. A greater sampling count implies higher search precision, as denser sampling can cover regions that sparser sampling might overlook. However, this simultaneously increases computational overhead.

To validate this theory, we employed Llama2-7B on 16k long texts, fixing two hyperparameters while adjusting one to compute the PPL values after applying the DCIS algorithm. As shown in Table \ref{tab:init_range}, we fixed $C=10$ and adjust the initial range $[l,r]$, and in Table \ref{tab:samp_cnt}, we fixed $[l,r]=[5,5]$ and adjust $C$. The experimental results are consistent with our analysis:

\begin{enumerate}
    \item Parameter robustness. The DCIS algorithm demonstrates excellent robustness to hyperparameter selection, where parameter variations within reasonable ranges do not significantly affect model performance.
    \item Practical guidance. In practical applications, one can select moderate initial ranges (e.g., $[-5,5]$) and the number of increments per segment (e.g., $C=10$) to ensure both search effectiveness and computational cost control.
    \item Algorithm stability. This parameter insensitive characteristic indicates that the DCIS algorithm possesses favorable stability and practicality.
\end{enumerate}

\begin{table}[t]
    \centering
    \resizebox{\columnwidth}{!}{
    \begin{tabular}{c c c}
    \toprule
        Metric & Initial Value & Value After DCIS Search \\
        \hline
        PPL & 19.25 & 10.19 \\
        \hline
        LongPPL & 12.99 & 2.64 \\
        \hline
    \end{tabular}
    }
    \caption{Comparison of different optimization objectives.}
    \label{tab:longppl}
\end{table}

\subsection{LongPPL Metric}

In our framework, PPL serves primarily as the optimization objective function for searching scaling factors. This design offers excellent flexibility, allowing for substitution with other evaluation metrics according to specific requirements, such as LongPPL \cite{fang2024wrong}. To further validate the scalability of the proposed framework, we adjusted the optimization objective from PPL to LongPPL and conducted comparative experiments based on the Llama2-7B model on 16k text sequences.

The experimental results presented in Table \ref{tab:longppl} demonstrate that employing LongPPL as the optimization objective can similarly yield superior scaling factors, with performance essentially consistent with using PPL as the optimization objective. This finding further confirms the excellent scalability of the framework proposed in this paper under different evaluation metrics, while also validating the effectiveness and robustness of the method.

\section{Conclusion}

In this paper, we have presented DCIS, a novel and efficient algorithm for identifying effective RoPE scaling factors to significantly enhance LLM length extrapolation. 
Our framework demonstrably mitigates performance degradation at extended target lengths. 
Crucially, DCIS enables models fine-tuned on shorter contexts to generalize effectively to considerably longer sequences, substantially reducing fine-tuning costs. 
Furthermore, the scaling factors identified by DCIS yield improved extrapolation performance even without any fine-tuning. 
Our investigation into non-strictly increasing scaling factors revealed their benefits for model performance. 
Comparative analyses underscore the effectiveness, efficiency, and robustness of DCIS, offering a practical solution for extending the operational context window of LLMs.

\clearpage
\section*{Acknowledgments}
The present research was supported by the National Key Research and Development Program of China (Grant No. 2023YFE0116400). We would like to thank the anonymous reviewers for their insightful comments.

\section*{Limitations}



We conducted a search for a scaling factors of 128k on a small model, Phi-3-mini-4k-instruct \cite{abdin2024phi}, with an actual context window size of 2k. We observe that the model's PPL remains around 70, with no significant decrease. Thus, such search algorithms appear to have certain requirements for the model's inherent capabilities and are unable to span too large a scaling factor at once. This also demonstrates that, relative to YaRN, which has a smaller scaling factor, the scaling factors identified through search have larger scaling factor, thereby more fully leveraging the model's extrapolative potential.

\bibliography{acl_latex}

\clearpage

\appendix

\section{Additional Results}

We present supplementary results obtained during our experimental process in this section.

\subsection{Fewer Fine-Tuning Steps} \label{apd:fts}

\begin{figure*}
    \centering
    \includegraphics[width=16cm]{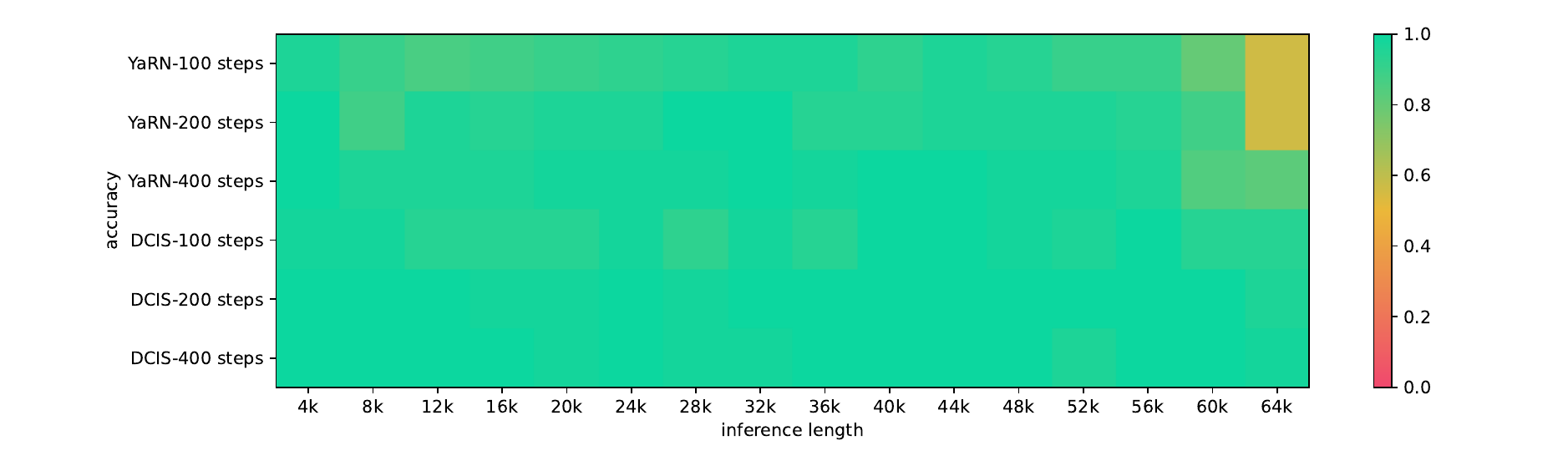}
    \caption{Comparison of the passkey results at different numbers of steps of fine-tuning on the 64k length.}
    \label{fig:passkey_t64k}
\end{figure*}

During the fine-tuning process of YaRN and DCIS, we observed that our method, DCIS, achieved PPL and passkey test results consistent with, or even superior to, those of YaRN after only 100 steps of fine-tuning, as depicted in Table \ref{tab:t64k} and Figure \ref{fig:passkey_t64k}. It is demonstrated that an initially superior scaling factors requires fewer fine-tuning steps.

\begin{table*}
    \centering
    \begin{tabular}
        {ccccccccccc}
        \toprule
        \multirow{2}{*}{{Method}} & Fine-tuning & \multicolumn{8}{c}{Evaluation Context Window Size} \\
        & Steps & 4k & 8k & 16k & 24k & 32k & 40k & 48k & 56k & 64k\\
        \hline
        \multirow{3}{*}{YaRN} & 100 & 3.71 & 3.53 & 3.01 & 2.78 & 2.68 & 2.60 & 2.54 & 2.48 & 2.45 \\
        & 200 & 3.71 & 3.53 & 3.01 & 2.78 & 2.68 & 2.60 & 2.54 & 2.48 & 2.44 \\
        & 400 & 3.71 & 3.52 & 3.00 & 2.78 & 2.67 & 2.59 & 2.53 & 2.47 & 2.44 \\
        \hline
        \multirow{3}{*}{DCIS(Ours)} & 100 & 3.72 & 3.53 & 3.00 & 2.77 & 2.67 & 2.59 & 2.53 & 2.47 & 2.43 \\
        & 200 & 3.72 & 3.53 & 3.01 & 2.77 & 2.67 & 2.59 & 2.53 & 2.47 & 2.43 \\
        & 400 & 3.71 & 3.52 & 3.00 & 2.77 & 2.66 & 2.58 & 2.52 & 2.46 & 2.42 \\
        \hline
    \end{tabular}
    \caption{Comparison of the PPL results at different numbers of steps of fine-tuning on the 64k length.}
    \label{tab:t64k}
\end{table*}

\subsection{Shorter Length of PPL without Fine-Tuning} \label{apd:slppl}

\begin{figure*}
    \centering
    \includegraphics[width=7cm]{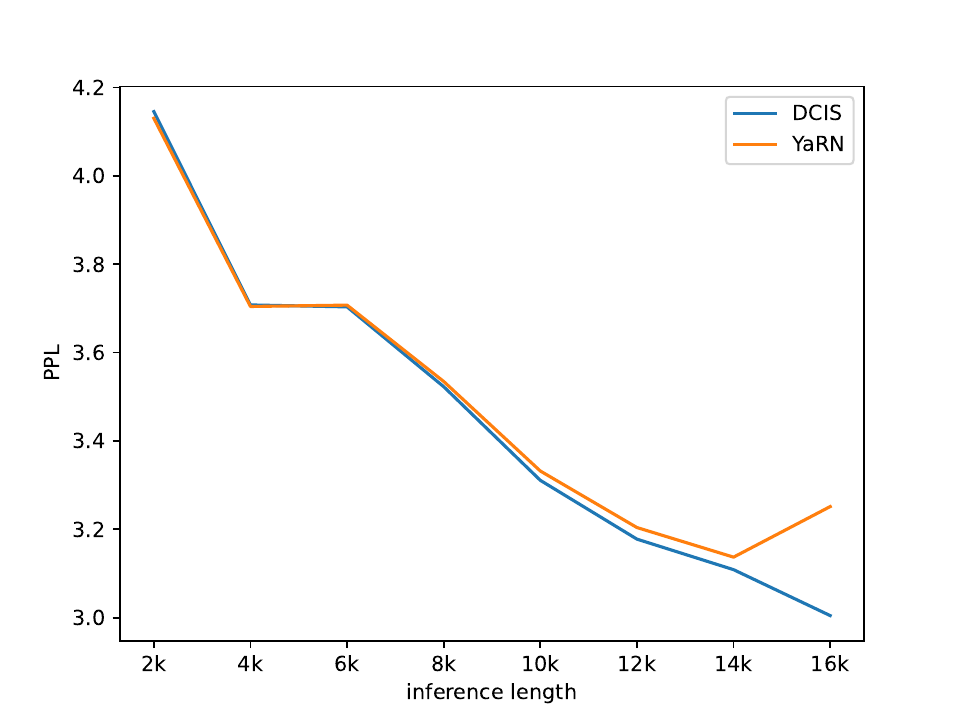}
    \includegraphics[width=7cm]{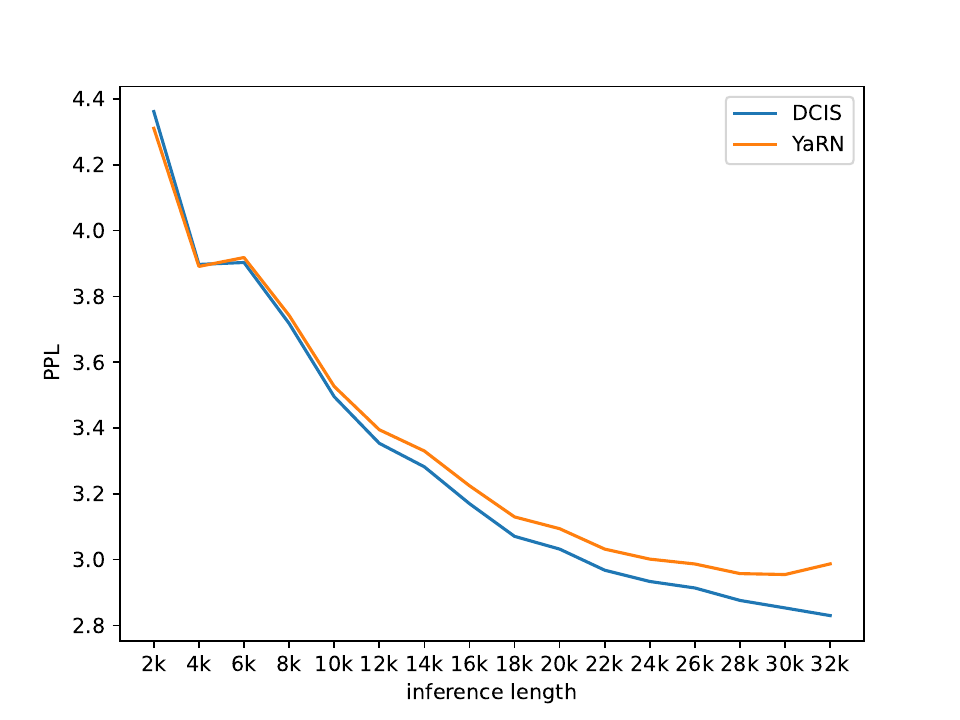}
    \caption{Left: 16k-length. Right: 32k-length.}
    \label{fig:nofts}
\end{figure*}

Even at shorter lengths of 16k and 32k, our scaling factors consistently achieved the lowest PPL, as illustrated in Figure \ref{fig:nofts}.

\begin{table*}[t]
    \centering
    \begin{tabular}
        {cccccc}
        \toprule
        \multirow{2}{*}{Method} & Fine-tuning & \multirow{2}{*}{ARC-c} & \multirow{2}{*}{Hellaswag} & \multirow{2}{*}{MMLU} & \multirow{2}{*}{TruthfulQA} \\
        & Length &  &  &  & \\
        \hline
        Original & - & 52.6 & 79.0 & 46.4 & 39.0 \\
        \hline
        \multirow{2}{*}{YaRN} & 64k & 52.8 & 78.8 & 42.1 & 39.0 \\
        & 16k & 52.6 & 78.4 & 42.4 & 38.4 \\
        \hline
        CLEX & 16k & 52.3 & 78.3 & 42.1 & 41.3 \\
        \hline
        \multirow{3}{*}{DCIS(Ours)} & 64k & 52.3 & 78.4 & 41.8 & 39.2 \\
        & 16k & 53.0 & 78.2 & 41.4 & 38.8 \\
        & 4k & 52.6 & 78.4 & 43.7 & 38.0 \\
        \hline
    \end{tabular}
    \caption{Evaluation of different methods on the benchmarks from the Hugging Face Open LLM Leaderboard}
    \label{tab:benchmarks}
\end{table*}

\subsection{Benchmarks} \label{apd:benchmarks}

As shown in Table \ref{tab:benchmarks} of the Open LLM Leaderboard, there is no clear superiority among different methods. Furthermore, the models with extended context windows do not exhibit significant performance degradation on short texts compared to their original counterparts.

\end{document}